# Look No Further: Adapting the Localization Sensory Window to the Temporal Characteristics of the Environment

Jake Bruce, Adam Jacobson and Michael Milford

*Abstract*—**Many localization algorithms use a spatiotemporal window of sensory information in order to recognize spatial locations, and the length of this window is often a sensitive parameter that must be tuned to the specifics of the application. This paper presents a general method for environment-driven variation of the length of the spatiotemporal window based on searching for the most significant localization hypothesis, to use as much context as is appropriate but no more. We evaluate this approach on benchmark datasets using visual and Wi-Fi sensor modalities and a variety of sensory comparison front-ends under in-order and out-of-order traversals of the environment. Our results show that the system greatly reduces the maximum distance traveled without localization compared to a fixed-length approach while achieving competitive localization accuracy, and our proposed method achieves this performance without deployment-time tuning.**

## I. INTRODUCTION

SENSORY information arrives in a temporal stream, with order and timing often being as important as content. When localizing against previous sensory data, which is an important capability for such varied domains as autonomous cars, unmanned aerial vehicles, and automated environmental and infrastructure monitoring, decisions must be made about the length of temporal window to consider and the right answer is highly dependent on sensory modalities and the environment. Temporal filtering approaches are known to be effective [1], and previous work has shown improvements in localization by matching windows of recent history against contiguous regions of recorded trajectories ([2], [3]), indicating that the spatiotemporal history of the agent is informative. Two of the main challenges in recognizing known locations are varation in environmental conditions, and changes in viewpoint. Although tackling these particular issues is an important direction for progress in the area, this paper addresses an orthogonal component of temporal algorithms: the question of how much history to consider [4].

JB, AJ and MM are with the Queensland University of Technology, and MM is with the ARC Centre of Excellence for Robotic Vision. jacob.bruce@hdr.qut.edu.au
*This work was supported by an Australian Research Council Future Fellowship FT140101229 to MM. JB is supported by an NSERC Postgraduate Scholarship and an ARC International Postgraduate Research Scholarship.

Digital Object Identifier (DOI): see top of this page.

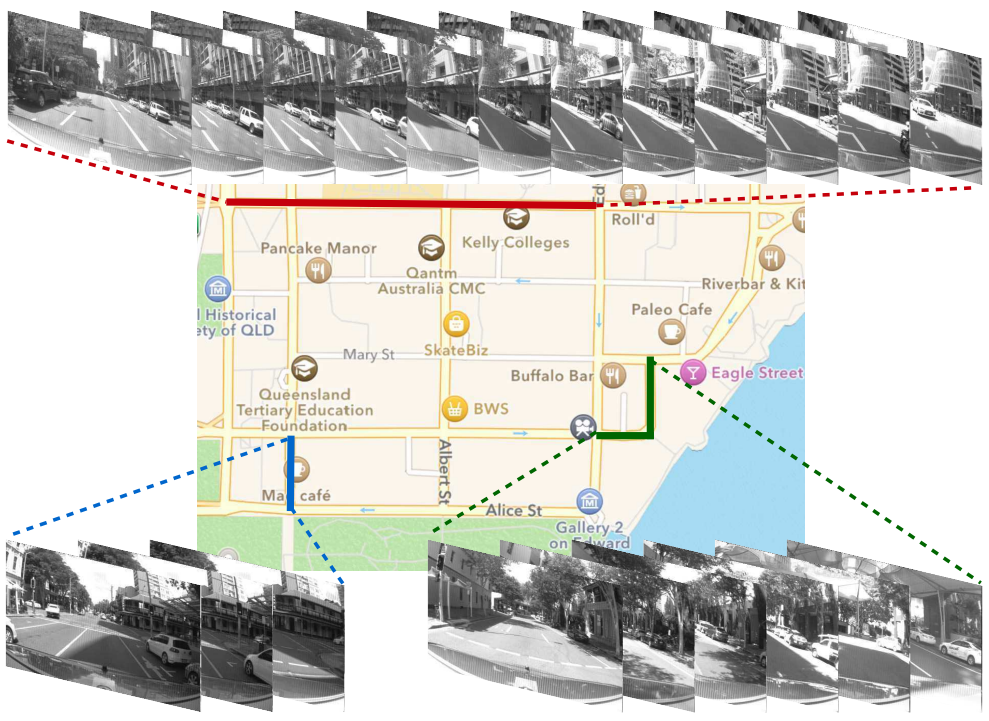

(a) Amount of relevant context varies between routes.

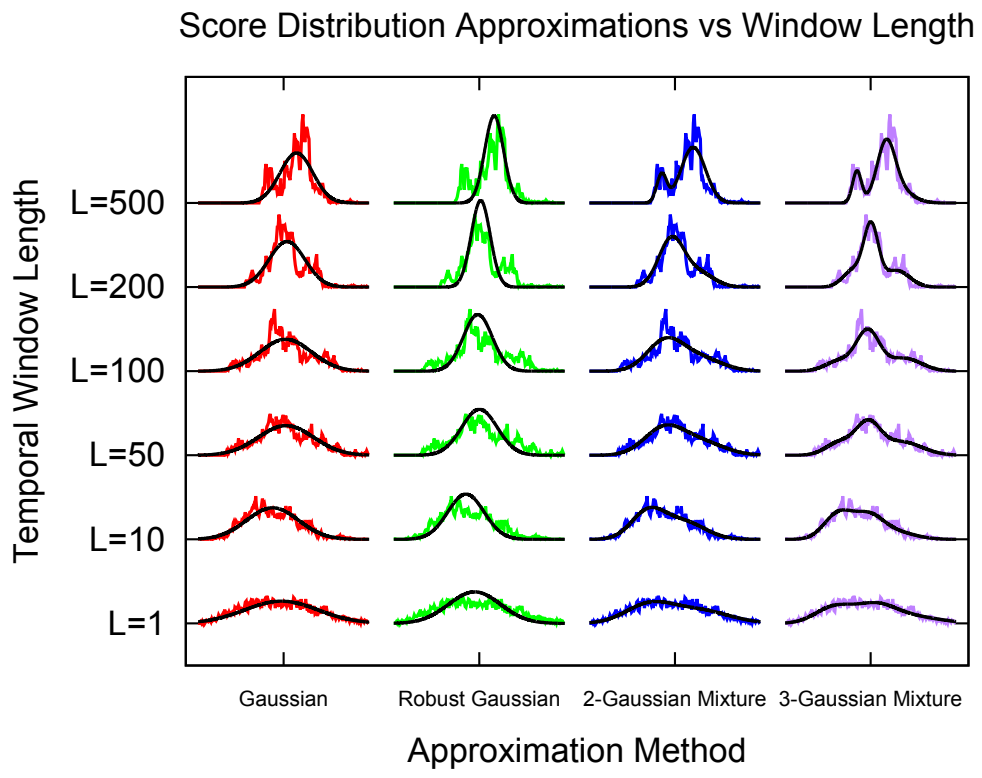

(b) Distribution of temporal window differences gets sharper with increasing window length $L$, and sharpness of fit varies with the approximation method.

Fig. 1: Variable window length can enable a localization system to consider the appropriate amount of temporal context, which can be determined by choosing a window length that produces the most significant localization hypothesis.

Temporal integration methods generally have some form of recency parameter, whether implicit or explicit, representing a temporal window that is often fixed in length over the entire traverse. Results are usually reported at multiple values of this parameter with the optimal length largely dependent on how densely the frames are spaced, which varies between datasets (*unknown frame density*). For well-behaved data where routes are traversed in the same order as the reference data, and in which sensory frames are recorded at consistent spatial separation, longer temporal windows tend to perform better ([5], [3], Fig. 1b). However, many practical cases involve routes that are traversed out of order such as the city blocks in Fig. 1a (*out-of-order traversals*) [6], and longer windows accrue drift in spatial separation that grows with the length, all else being equal (*varying frame separation*) [7].

To summarize, three characteristics that negatively impact the performance of fixed-length window approaches are *unknown frame density*, *out-of-order traversals*, and *varying frame separation*. In addition to these concerns, using longer temporal windows generally incurs longer periods of uncertainty before achieving a high-confidence position estimate, and localization latency is an important metric to quantify the degree of reliance on open-loop state estimation ([8], [9]).

Without prior knowledge of the environment it is not possible to predetermine a window length that will succeed in all cases, and no techniques have yet been developed for determining this window length online. In this work, we propose a method to vary the temporal window length to consider over the course of a traversal, with the goal of using as much temporal context at each point as is necessary to produce the strongest position hypothesis, and no more. Unlike the particle filter approach in [6], our method requires no knowledge of the topology of the road network, and is suitable for online localization in contrast to offline batch methods such as [10]. We show that our adaptive window achieves much lower localization delays on in-order and shuffled versions of three benchmark datasets due to opportunistic use of short windows while maintaining accuracy competitive with a fixed length approach, with no prior knowledge of road network topology or environmental characteristics. The performance of the proposed approach is also less sensitive to parameter choices than the baseline approach on all benchmarks, requiring little to no application-specific tuning to meet and even exceed the performance of the baseline approach.

## II. RELATED WORK

In this section we review current work in spatiotemporal localization and the decisions that have been made regarding the amount of temporal history to take into consideration.

Single-frame approaches for localization in a sensory stream attempt to find a match in a database for each incoming frame, and this has been demonstrated using image description techniques such as global "gist" representations [11], feature vocabularies ([12], [13]), and neural networks [14]. The sequential nature of sensory experience is considered implicitly by approaches such as particle filters [1], flow networks [15], Markov models [16], experience-based mapping [17], and continuous attractors [18] which converge on location estimates as evidence accumulates over time. Temporal methods generally have implicit or explicit parameters affecting the amount of recent context to consider, such as particle resampling probabilities, flow network connectivity constraints, match quality thresholds, or hypothesis decay rates.

Sequence-based localization approaches explicitly consider the temporal contiguity of sensory streams to disambiguate otherwise similar location hypotheses using context ([2], [3], [10], [19]), and this has been shown to produce reasonable performance even with very low sensor quality [20]. These sequence-based approaches generally use a fixed-length temporal window for comparison against reference trajectories ([10] being a notable exception in which contiguous matches are identified regardless of size, although it is an offline batch-processing approach and has not yet been generalized to the online localization problem). All else being equal, longer temporal windows are less likely to perfectly overlap due to speed variation, and mitigating this property by explicitly sampling based on distance has been shown to improve performance [7].

Grouping sequences of images together for recognizing places between changing conditions has been shown to improve computation time as well as performance [21]. Similar in concept to our proposal for adaptive window lengths but tackling a different problem, scalable methods for large location databases have been proposed to avoid searching the entire localization history of the agent at every point [4]. Unsupervised learning has been used to produce environment-driven transformations of navigation imagery to improve sequence-based approaches in the face of appearance change ([22], [23]), and empirical distributions over whole-image descriptor differences have been used to provide probabilistic interpretations of single-image match scores [24]. To cope with loops and other out-of-order traversal characteristics, spatial information such as GPS priors have been shown to improve sequence-based localization [25]. Particle filters have been combined with SeqSLAM to improve computational efficiency [26]. [6] uses a particle filter to improve accuracy and reduce time to localization on out-of-order traversals of intersecting regions such as city blocks, but this differs from our approach in that it requires prior knowledge of the topology of the road network.

Although temporal methods cannot avoid making decisions about how much spatiotemporal history to consider, we propose an approach to make this decision explicit, dynamic, and environment-driven, by searching a range of temporal window lengths to produce the most significant localization hypothesis. This allows a spatiotemporal matching system to use as little history as possible to produce the best hypothesis, but no less.

## III. APPROACH

The proposed method is an environment-driven approach for adaptively varying the amount of spatiotemporal context to consider. In algorithms such as SeqSLAM [2] and SMART [3], a fixed-length window of recent sensory data is matched against stored observations in order to localize a mobile agent. Two important challenges in recognizing known locations are varation in environmental conditions, and changes in viewpoint; however, this paper addresses an orthogonal issue: determining the length of the window that is compared against stored experience. We use SeqSLAM to illustrate our adaptive method, although the mechanism of varying the window length in response to the observed hypothesis score distribution is applicable to any approach that attempts to match spatiotemporal windows in sensory streams such as SMART [3] and dynamic time warping [27]. The approach proceeds as follows.

## A. SeqSLAM

In SeqSLAM, a temporal window of recent sensory data is compared against a reference traversal that constitutes a topological map of the environment; the algorithm consists of a per-image frame comparison, and a window-based sequence matching step. First, a difference metric $d_{ij}$ is computed between each query frame $Q_i$ and each frame $R_j$ in the reference traversal (treating each image as i.i.d. as in [2]), and these differences are normalized per $Q_i$ to account for varying frame intensity:

$$d_{ij} = \frac{|Q_i - R_j| - \mu_{di}}{\sigma_{di}}$$

where $\mu_{di}$ and $\sigma_{di}$ are the mean and standard deviation, respectively, of $|Q_i - R_j|$ over the set of all reference frames $R_j$ from previous traversals. We use sum of absolute differences (SAD) as the difference operator here for the purpose of demonstration, although the method applies to other difference metrics as described in the experimental section.

Given a collection of difference values $d_{ij}$, the recent history of length $L$ is compared against all positions in the reference trajectory by computing the total difference $s_{ij}(L)$ between the most recent $L$ sensory observations and all temporal windows of length $L$ in the reference trajectory, normalized by length:

$$s_{ij}(L) = \frac{1}{L}\sum_{k=0}^{L-1} d_{i-k,j-k}$$

which is then used as the localization score for each potential location to match against, and the reference location $j$ that minimizes $s_{ij}(L)$ is considered the most likely localization hypothesis at query location $i$.

## B. Adaptive Window Length

SeqSLAM implementations to date have used a fixed value for $L$ for each entire traversal. However, the position hypothesis score $s_{ij}(L)$ depends on the window length parameter $L$, so it is instructive to investigate how the distribution of $s_{ij}(L)$ varies with $L$.

At $L = 1$, the distribution of $s_{ij}(L)$ is equivalent to the distribution of single-frame differences $d_{ij}$. Fig. 1b shows the distribution of history differences as a function of $L$, with the distribution generally becoming narrower as $L$ increases. This reinforces the intuition that longer window lengths are more discriminative, all else being equal. Importantly, if a strong match for the current location exists in the reference trajectory, this will tend to be a significant outlier on the low end of the history difference distribution, and this forms the basis for our method of adapting the window length $L$.

As shown in Fig. 1b, history differences tend to be distributed more narrowly as $L$ increases. To determine the value of $L$ that produces the strongest position hypothesis, we first fit an approximation to the distribution of $s_{ij}(L)$. In this work, we assume the distribution of window differences is sufficiently Gaussian, and compare four different methods for approximating the score distribution: a normal distribution, a median-based robust normal distribution, and Gaussian mixture models with 2 and 3 components; these are computed as follows.

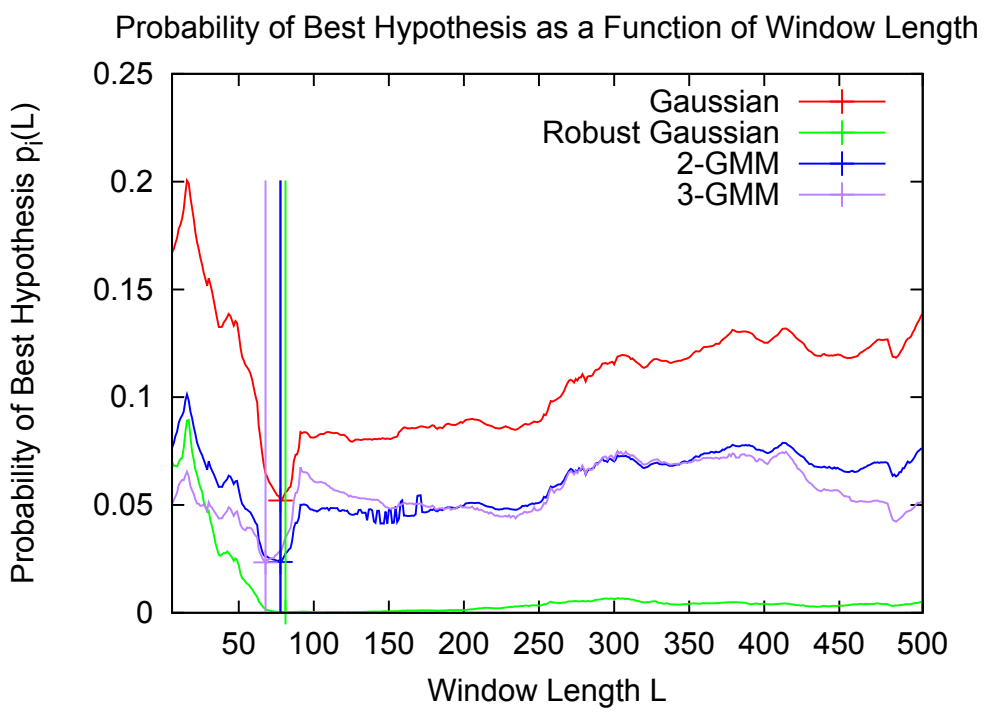

Fig. 2: $p_i(L)$ as a function of $L$ at a frame near the middle of the Eynsham dataset computed with respect to different distribution approximations, with the global minimum of each curve indicated by a vertical line. Note that the curves for the different approximation methods have qualitative similarities, and each yields a minimum at a similar $L$.

*Gaussian* – a normal distribution $\mathcal{N}(\mu_i, \sigma_i)$, with location and scale parameters $\mu_i$ and $\sigma_i$ set to the mean and standard deviation of window differences $s_{ij}(L)$ between every location in the reference traversal and the query location $i$.

*Robust Gaussian* – an outlier-robust version of the normal distribution, given by $\mathcal{N}(\text{median}_i, k \cdot \text{MAD}_i)$ with location and scale parameters $\text{median}_i$ and $\text{MAD}_i$ set respectively to the median and median-absolute-deviation:

$$k \cdot \text{MAD}_i = k \cdot \text{median}(|\text{median}_i - s_{ij}(L)|)$$

for all $s_{ij}(L)$ between the frames in the reference traversal and the current query frame $i$, with constant scale factor $k \approx 1.4826$ for normally distributed data [28].

*Gaussian Mixture Model* – a mixture model of $K$ Gaussian components fit using 10 iterations of expectation-maximization using the mean and standard deviation as with the single Gaussian model above. In this work, we compare two different mixture models, using $K = 2$ and $K = 3$ Gaussian components.

Given an approximation of the distribution over the scores of each possible temporal window, we then compute $p_i(L)$, the instantaneous probability of drawing the best location hypothesis from the approximated distribution. A representative example of the behavior of $p_i(L)$ with respect to $L$ is shown in Fig. 2. We then choose the window length $L_i$ to use for localization at frame $i$ that minimizes $p_i(L)$—informally, we find the length that produces the most significant hypothesis:

$$L_i = \operatorname*{argmin}_L p_i(L)$$

In this work we use the range $10 \le L \le 500$ to be on equal terms with the range of the fixed windows against which we compare results.

Over the course of a traversal, the amount of history used by the system will vary based on how much context is necessary to narrow down the distribution of position hypothesis scores. Fig. 3 shows the chosen window length $L$ over the course of the Wi-Fi dataset, demonstrating that the amount of history required to narrow down the position estimate varies throughout the traverse, and the adaptive method tends to choose shorter window lengths in the shuffled traverse.

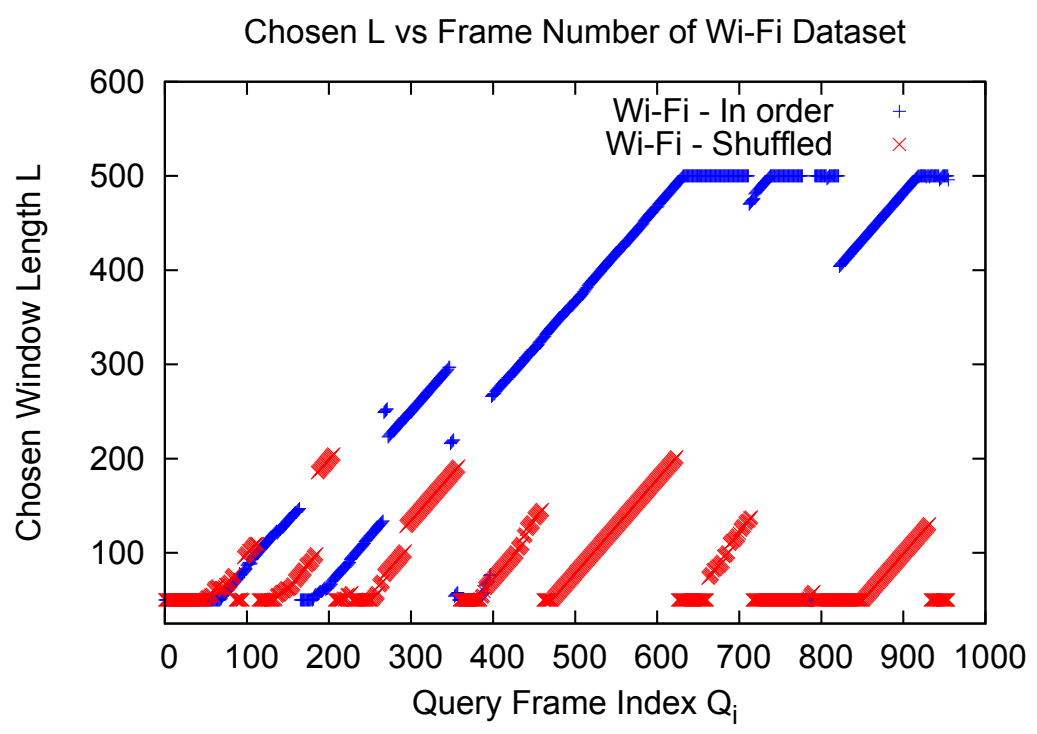

Fig. 3: Chosen window length varies over the course of navigation, shown here on the Wi-Fi dataset with $L$ for the in-order traversal shown in blue, and in red for the shuffled traversal. Note that the proposed adaptive method tends to choose shorter windows in the shuffled version as expected.

## IV. EXPERIMENTAL SETUP

We evaluated the method on a set of navigation datasets each with its own distinctive characteristics. To reiterate, the main challenges addressed by adjusting the length of the temporal window are *unknown frame density*, *out-of-order traversals*, and *varying frame separation*. The benchmark datasets evaluated for this paper exhibit the specific challenges the method is proposed to address.

### *Eynsham dataset*

The first dataset on which we evaluate is the Eynsham dataset [29], a car dataset of panoramic imagery consisting of a 35km route and in-order traversal with median frame separation distance of 6.7m, but this distance is imperfect because it was enforced by GPS estimates. The difference operators used for this dataset were the sum of absolute differences (SAD) on pairs of 32x16-pixel downsampled greyscale images as used by [20][1], and the cosine difference of the feature activations of the pre-final layer of a state-of-the-art convolutional neural network (CNN) trained on a scene recognition task [30] (an approach similar to [31]). We use these different sensory comparison front-ends to demonstrate that the proposed adaptive window length system can generalize to a variety of sensory comparison techniques. Fig. 5 shows a map of the route and a sample pair of frames matched with our approach.

### *CBD dataset*

Next we present results on a panoramic image dataset captured in a car traversing the city blocks of the Brisbane Central Business District (CBD) [6]. The CBD dataset consists of a 3.6 km reference traverse during the and a 6 km naturally out-of-order query traverse at night, with panoramic images taken at odometry-enforced one-meter intervals. The difference operator for this dataset is designed for challenging day-to-night appearance change between traversals, and involves a SAD operator on 360x32-pixel grayscale images with contrast enhancement, sky blackening and patch normalization as described in [6]. In addition to evaluating the value of the adaptive window length for out-of-order traversals, this dataset provides another opportunity to show that the proposed method can be applied on top of complex image comparison techniques. Fig. 6 shows a map of the reference and query routes and an example frame pair matched with our approach.

### *Campus Wi-Fi dataset*

To investigate the ability of the adaptive method to generalize to different sensing modalities in addition to inconsistent frame separation, we also evaluated localization using only wireless access point fingerprints. The campus Wi-Fi dataset consists of a human traversing a campus environment on foot while recording the MAC address and signal strength of all visible wireless access points. During the course of the reference and query traversals 709 unique access points were observed with 12.6 present per frame on average, so we construct a sparse sensory vector for each frame, spaced one second apart at a variable walking pace of approximately one meter per second. The sparsity and spatial characteristics of Wi-Fi access point signal strength vectors are very different from images (see the sparse vectors in Fig. 7) so this also illustrates the ability of the system to cope with a variety of spatial sensing characteristics without sensor-specific tuning.

### *Data processing*

We compared the proposed adaptive method against the fixed-length approach with a range of window lengths $L \in \{10, 25, 50, 100, 200, 350, 500\}$. The adaptive approach was evaluated with values of $L_{min}$ and $L_{max}$ of 10 and 500 respectively, to compare on equal terms with the range of window lengths chosen for the fixed-length approach.

Since the Eynsham and Wi-Fi datasets consist of in-order traversals, we also present results on synthetically shuffled versions of these datasets created by randomly rearranging the order of contiguous segments between 2% and 20% of the length of each dataset for the query traverse. The ground truth frame correspondences for each dataset in original and shuffled order (where appropriate) are shown in Fig. 4. We would like to stress that this artificial rearrangement is included to demonstrate the sensitivity of the fixed window approach to out-of-order traversals: the CBD dataset consists of a naturally out-of-order traverse of a real road network. Image datasets were patch-normalized by partitioning each image into a grid of non-overlapping 4x4-pixel patches and normalizing each patch by subtracting its mean and dividing by its standard deviation, as described in [2]. The Wi-Fi dataset used raw access point signal strength vectors with no analogous normalization.

[1]As demonstrated in [20], aggressively downsampled imagery can still result in successful localization with SeqSLAM due to the relative insensitivity of low-resolution images to imprecise positioning and changing conditions, and the process of integrating sensory observations over time.

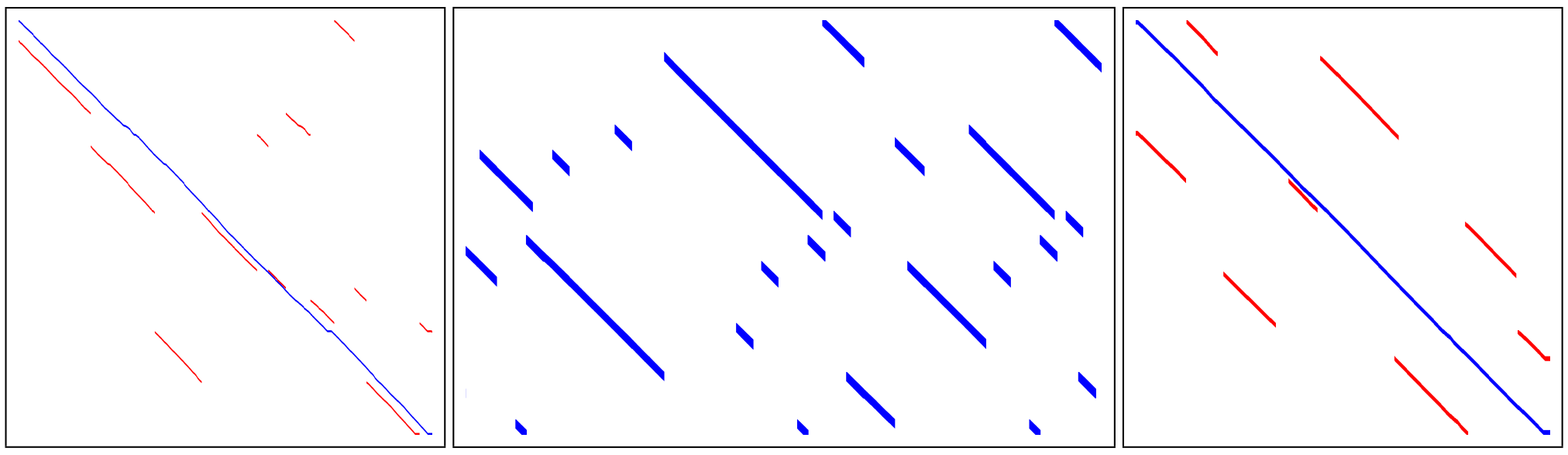

(a) Eynsham dataset (b) CBD dataset (c) Wi-Fi dataset

Fig. 4: Ground truth frame correspondence for each dataset, with query frame index along the horizontal axis and reference frame index along the vertical axis. Original versions of the datasets are shown in blue, and shuffled versions of the Eynsham and Wi-Fi dataset are in red.

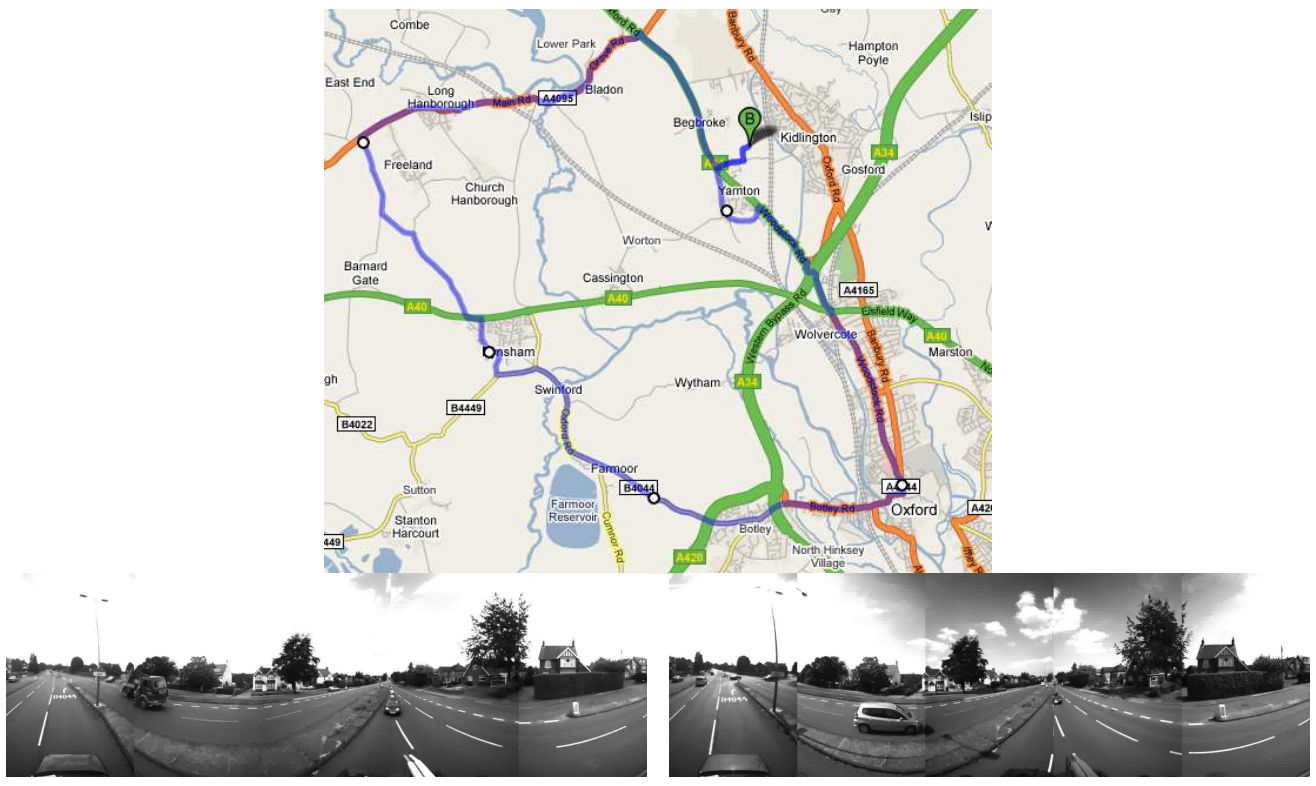

Fig. 5: Eynsham dataset map, with original query set being an in-order repeat traverse of the reference set, and an example image pair matched with our approach.

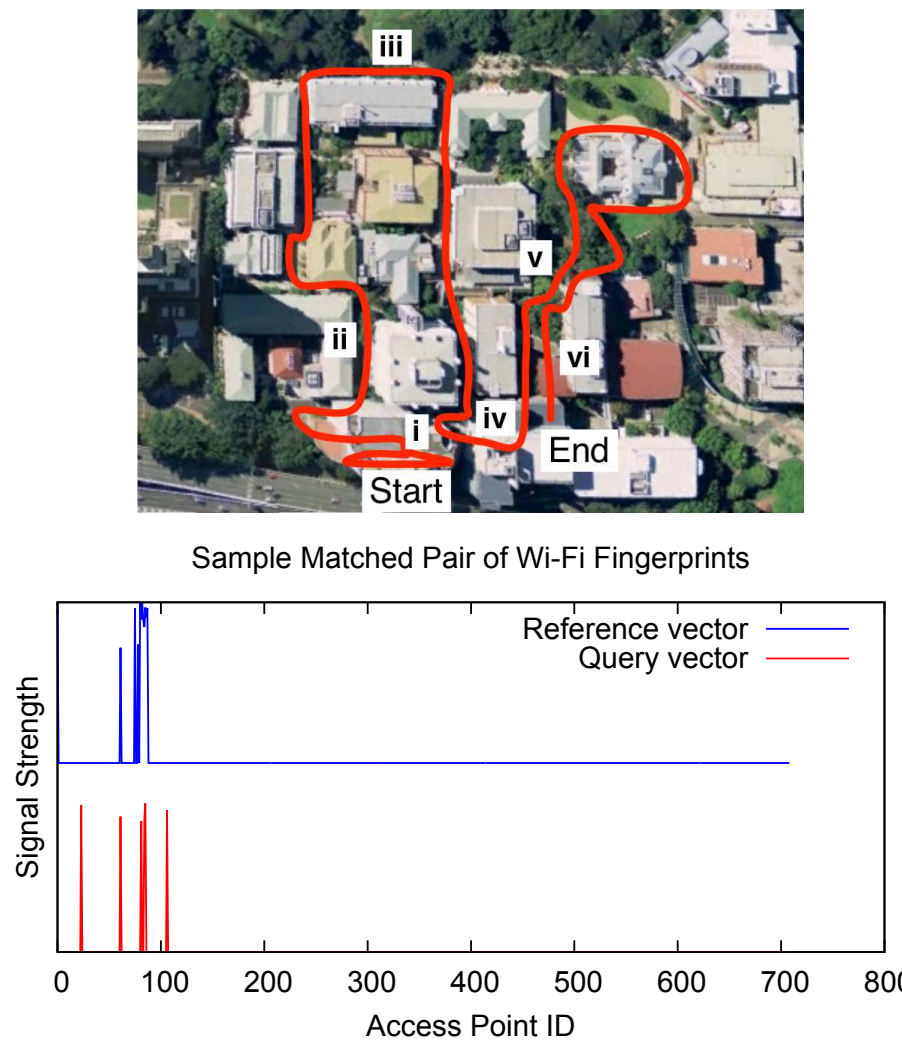

Fig. 7: Repeated trajectory in the Campus Wi-Fi dataset, and a pair of wireless fingerprints matched by our approach.

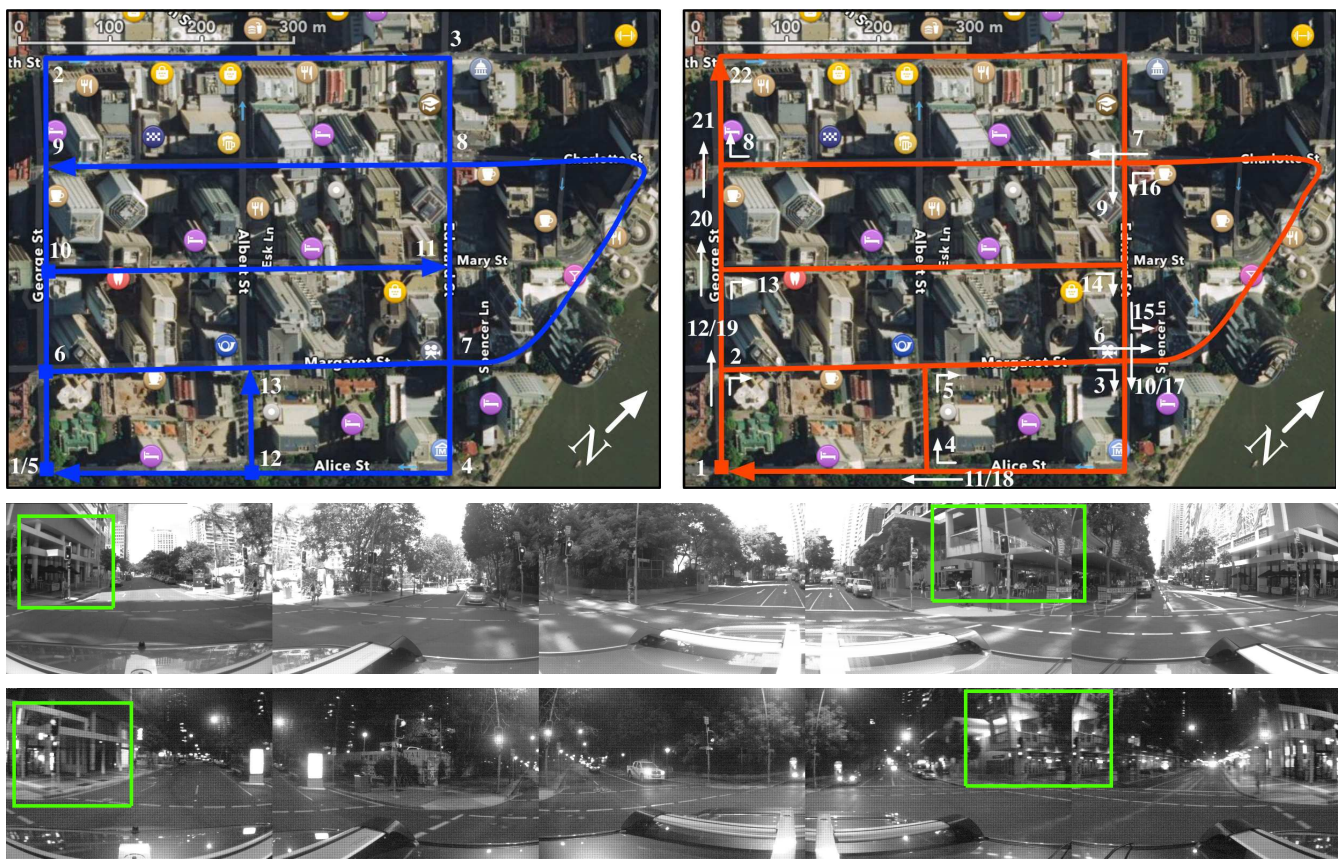

Fig. 6: CBD dataset map with reference and query trajectories illustrated in blue and red respectively, intersections numbered in the order they were traversed [6], and a sample pair of frames matched with our approach. Note that matches between day and night can be difficult to identify by eye, so we have highlighted structural cues in green as an aid to the reader.

## V. RESULTS

We evaluated the methods quantitatively using the following metrics on each dataset:

- *Maximum time-to-localization* (MTL), or "open-loop distance" ([8], [6], [9]), which is the maximum number of consecutive frames per traverse for which the system did not produce a correct location match.
- *Area under the precision-recall curve* (AUC) [32], which measures the accuracy of the system with respect to a search over match thresholds, where an AUC value of 1 indicates perfect performance at all thresholds.

The primary evaluation metric for the proposed adaptive method is the *maximum time-to-localization*: note the difference in scale of the vertical axis between figures. Figs. 8a and 9a show the time to localization on the original and shuffled versions of the Eynsham dataset, in which the adaptive method achieved maximum latencies equivalent to the shortest fixed windows using the Gaussian mixture model approximation on the original ordering, and all approximation methods were equivalent on the shuffled ordering. On the Wi-Fi dataset as shown in Figs. 8b and 9b, the adaptive method achieved equivalent latencies to the best fixed windows using the Gaussian and robust Gaussian approximations on the original ordering and outperformed all fixed windows on the shuffled ordering. On the naturally out-of-order CBD dataset shown in Fig. 8c, the adaptive window methods also achieved equivalent latencies to the best fixed windows, with the robust Gaussian approximation achieving lower latency than all fixed windows.

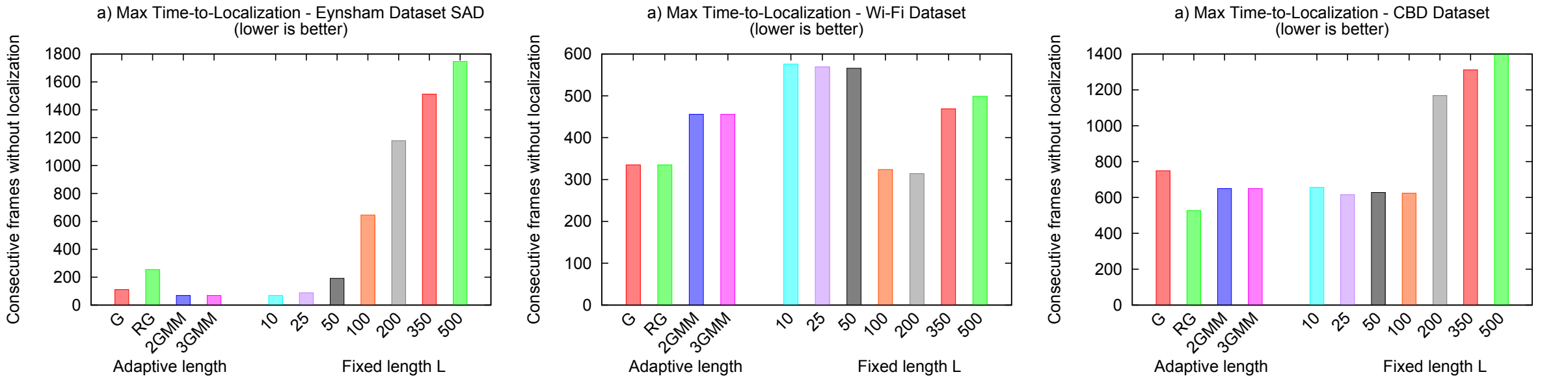

Fig. 8: MTL metric for a) Eynsham, b) Wi-Fi, and c) CBD

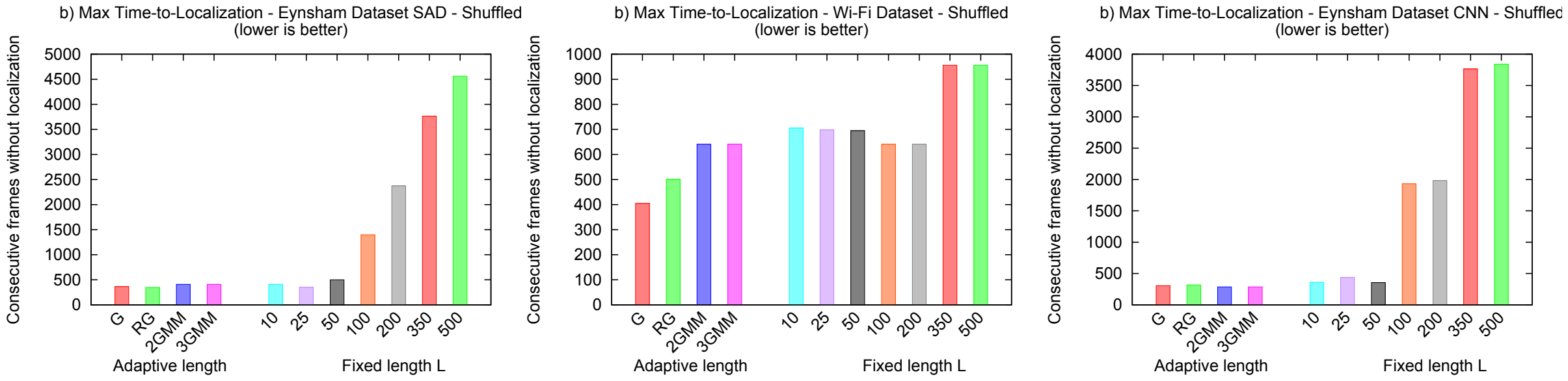

Fig. 9: MTL metric for shuffled versions of a) Eynsham, b) Wi-Fi, and c) Eynsham CNN

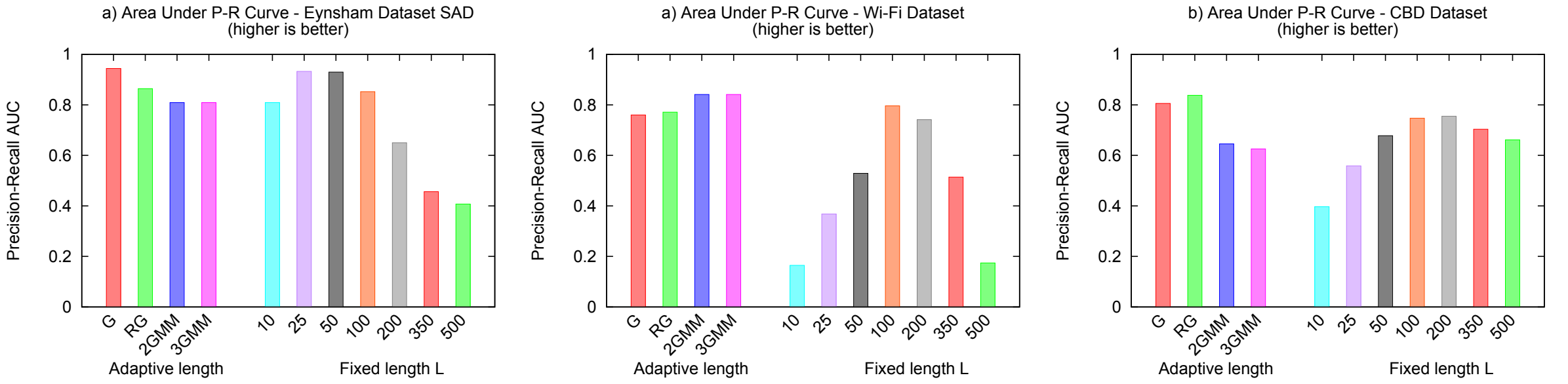

Fig. 10: AUC metric for a) Eynsham, b) Wi-Fi, and c) CBD

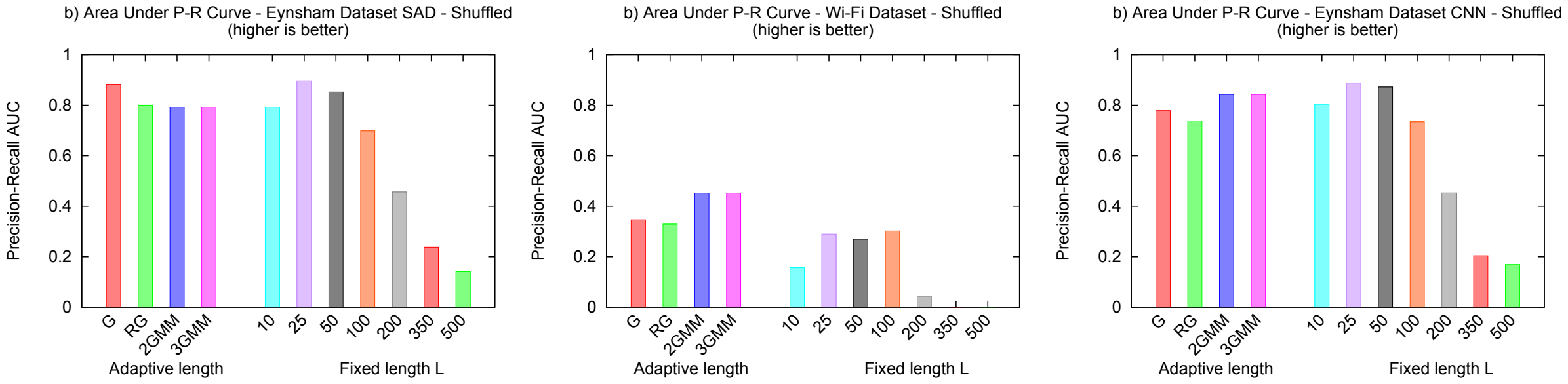

Fig. 11: AUC metric for a) Eynsham, b) Wi-Fi, and c) Eynsham CNN

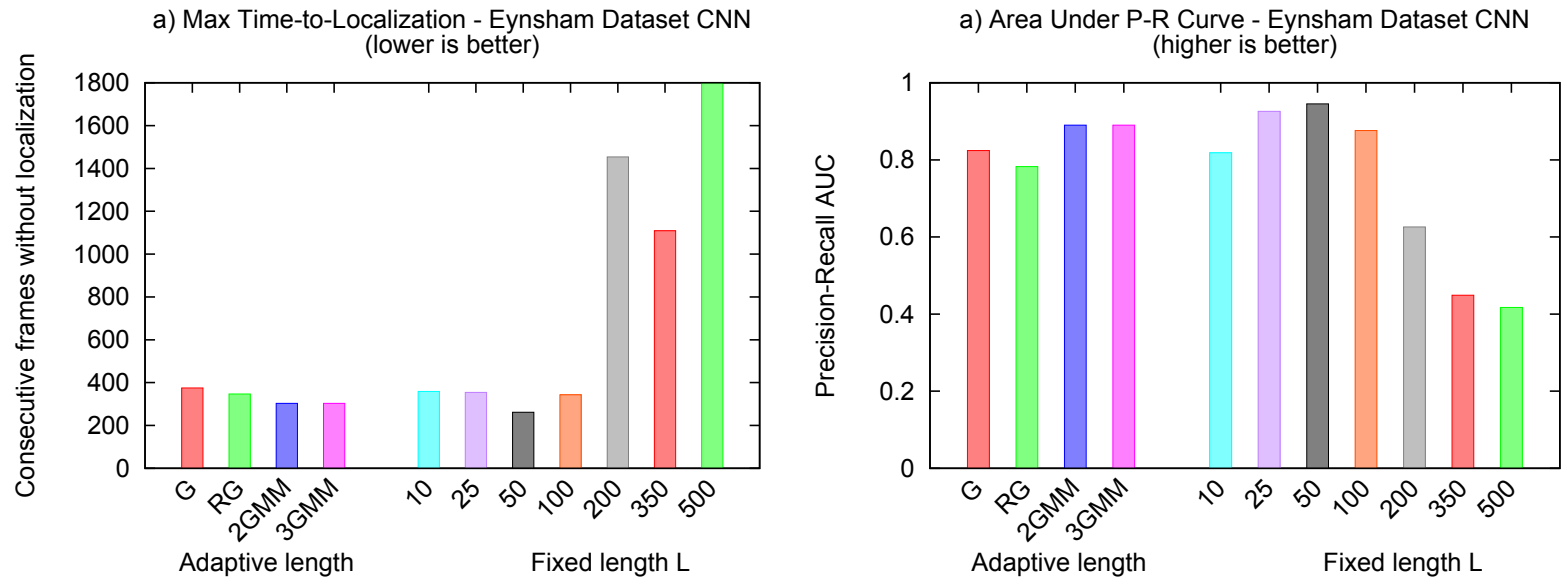

Fig. 12: Eynsham CNN a) MTL metric and b) AUC metric

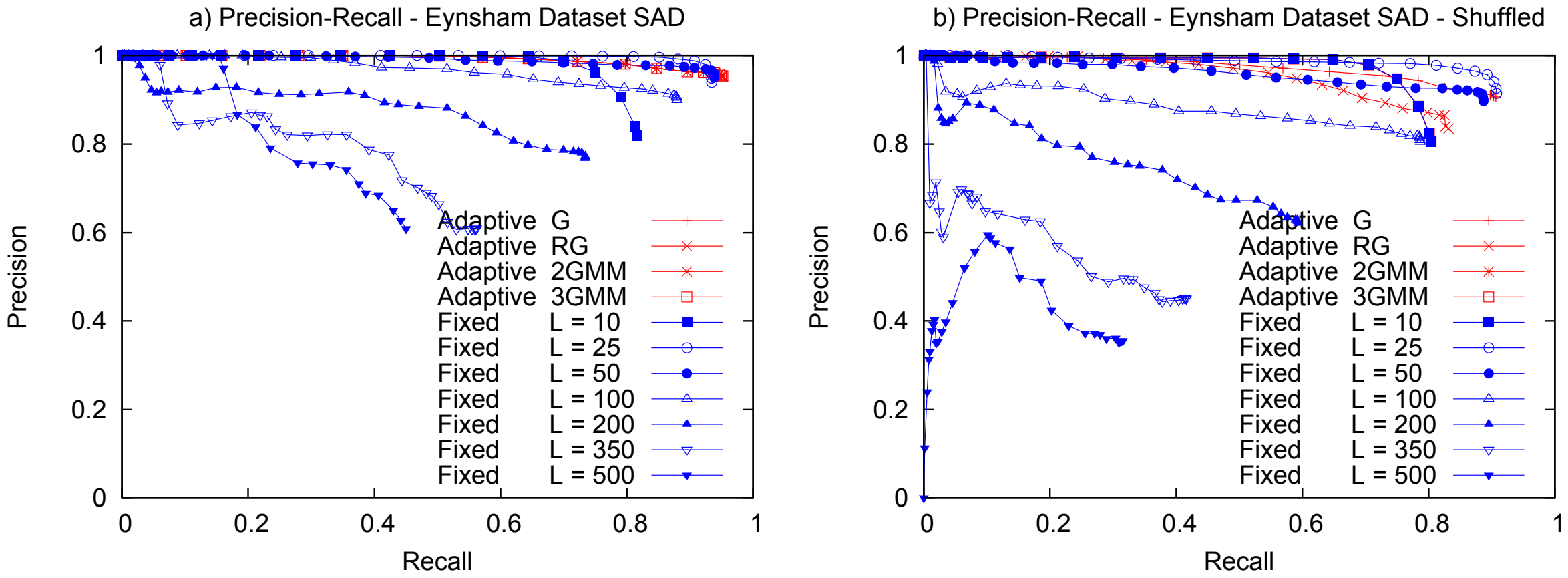

Fig. 13: Precision-Recall curves for a) original and b) shuffled Eynsham dataset, produced by varying the acceptance threshold on history difference scores. The fixed-window performance depends heavily on the length parameter.

The secondary evaluation metric is the *area under the precision-recall curve*. Fig. 10b shows the AUC results on the CBD dataset, in which the adaptive method using the Gaussian and robust Gaussian approximations achieved greater accuracy than the fixed-length methods. In both the original and shuffled orderings of the Eynsham dataset shown in Figs. 10a and 11a, the proposed adaptive methods achieve equivalent accuracy to the best fixed windows, with longer fixed windows performing significantly worse than the proposed approach. For the Wi-Fi dataset shown in Figs. 10b and 11b, our adaptive approach produced greater accuracy than the fixed window in both orderings, with the exception of length 100 on the original ordering, which achieved similar accuracy to our method.

We report additional results comparing our method using CNN feature activations against the fixed length approach in Figs. 9c, 11c and 12 in order to demonstrate that the results with the pixelwise SAD operator generalize to other sensory comparison front-ends. In Figs. 9c and 12a, our adaptive method achieves latencies equivalent to the best fixed windows on both orderings of the Eynsham dataset using CNN features, and Figs. 11c and 12b show that the adaptive method is almost as accurate as the best fixed windows, performing significantly better than the longest fixed windows.

The results show that the proposed adaptive approach is less sensitive than the fixed-length method in terms of both localization latency (MTL) and accuracy (AUC) to the ordering and sensory modalities of the datasets. Although the four distribution approximation methods (Gaussian, robust Gaussian, and mixture models with 2 and 3 Gaussian components) varied slightly in latency and accuracy, the choice of approximation had a relatively small effect on the system, and therefore the simple and computationally inexpensive Gaussian approximation is a sensible default. This is in contrast to the fixed length parameter $L$, for which different values produce large changes in performance depending on the dataset and traversal ordering. In particular, the fixed window AUC results exhibit a shape over different window lengths with a clear peak around $L = 200$ for the CBD dataset, $L = 25$ for Eynsham, and $L = 100$ for the Wi-Fi dataset; this shows that the ideal window length is dependent on the specifics of the environment and the agent, and this justifies our proposal of an adaptive method that is less sensitive to these variations. Fig. 13 shows this in the form of a Precision-Recall curve, where different values of $L$ result in significant differences to the curve, while varying the score distribution approximation has comparatively little effect.

## VI. DISCUSSION

The main goal of the proposed adaptive method is to use shorter window lengths where appropriate in order to reduce the maximum open-loop distance on each traverse. For all the datasets evaluated in this work, the adaptive method performs much better than the fixed-length approach on average, and performance is consistent between different choices of score distribution approximation. The naturally out-of-order CBD dataset and the synthetically rearranged versions of the Eynsham and Wi-Fi dataset provide additional justification for the adaptive method due to its ability to choose shorter window lengths in response to segments encountered out-of-order as demonstrated in the shuffled Wi-Fi dataset in Fig. 3. With respect to localization accuracy as measured by the area under the precision-recall curve, the adaptive method performs at levels competitive with and in most cases better than the fixed-length approach for all parameter values.

Although our results show that variation in history length improves performance on both metrics under the conditions we've evaluated, the technique is not appropriate in every situation. For example, datasets such as train routes [5] may show little benefit from the adaptive method, due to low variation in frame separation and no possibility for out-of-order traversal. Many other applications do exhibit such variations however, and the proposed approach is widely applicable to localization problems in challenging domains including autonomous cars, aerial vehicles, and industrial applications such as mining and automated inspection.

This paper proposes a novel method for adaptively varying the amount of spatiotemporal information used to compare the recent history of a mobile agent against a reference trajectory, and we have shown that the extension can dramatically reduce time to localization and maintain competitive accuracy, while requiring no prior knowledge of the environment. The technique can also apply to other spatiotemporal techniques for handling difficult localization problems such as dynamic time warping [27], and generalizes to a variety of sensory comparison front-ends including state-of-the-art CNN feature activations, having the potential to scale with further improvements to sensory processing front-ends in the future.